\documentclass[runningheads]{llncs}
\usepackage{graphicx}
\usepackage{enumerate}
\usepackage{wrapfig}
\usepackage{multirow}
\usepackage[table]{xcolor}
\usepackage{boldline} 
\usepackage{subcaption}
\usepackage[misc]{ifsym}
\usepackage{array}
\usepackage{booktabs}
\usepackage{hyperref}

\begin{document}
\title{Short-term Renewable Energy Forecasting in Greece using Prophet Decomposition and Tree-based Ensembles}
\titlerunning{Short-term RES forecasting: A Greek use case}
%
\author{Argyrios Vartholomaios\inst{1}\orcidID{0000-0001-8522-3550} \and
Stamatis Karlos\inst{1}\orcidID{0000-0002-5307-6186} \and  Eleftherios Kouloumpris\inst{2}\orcidID{0000-0003-1214-3845}
Grigorios Tsoumakas\inst{1,2}\orcidID{0000-0002-7879-669X}}
\authorrunning{A. Vartholomaios et al.}
%
\institute{School of Informatics, Aristotle University, 54124 Thessaloniki, Greece\\
\email{\{asvartho,stkarlos,greg\}@csd.auth.gr}\\ \and 
Medoid AI, 130 Egnatia St., 54622 Thessaloniki, Greece\\
\email{\{greg,lefteris.kouloubris@medoid.ai\}@medoid.ai}\\
}

\maketitle              

\begin{abstract}
Energy production using renewable sources exhibits inherent uncertainties due to their intermittent nature. Nevertheless, the unified European energy market promotes the increasing penetration of renewable energy sources (RES) by the regional energy system operators. Consequently, RES forecasting can assist in the integration of these volatile energy sources, since it leads to higher reliability and reduced ancillary operational costs for power systems. This paper presents a new dataset for solar and wind energy generation forecast in Greece and introduces a feature engineering pipeline that enriches the dimensional space of the dataset. In addition, we propose a novel method that utilizes the innovative Prophet model, an end-to-end forecasting tool that considers several kinds of nonlinear trends in decomposing the energy time series before a tree-based ensemble provides short-term predictions. The performance of the system is measured through representative evaluation metrics, and by estimating the model's generalization under an industry-provided scheme of absolute error thresholds. The proposed hybrid model competes with baseline persistence models, tree-based regression ensembles, and the Prophet model, managing to outperform them, presenting both lower error rates and more favorable error distribution.

\keywords{Time series forecasting  \and renewable energy sources \and signal decomposition \and Prophet model \and tree-based ensembles.}
\end{abstract}

\setcounter{footnote}{0} 
%
%
\section{Introduction}

Having achieved the liberalization of the energy market, Greece is working towards the development and operation of a competitive and economically viable energy model, through its admission into the unified European energy market. This transformation is guided by international directives that encourage the increased penetration of renewable energy sources (RES) into the local energy grid. Accurate RES forecasting is of utmost importance for transmission system operators (grid managers), in order to orchestrate the injection of RES into a robust energy grid. 

In the energy generation domain, short-term forecasting ranges from one hour or one day to one week, and is used in regulating the energy market, determining energy imports and exports, and arbitrating energy prices. Whereas long-term forecasting ranges from one to multiple years, and it serves the grid manager in long-term planning in relation to maintenance and expansion of infrastructure, investments, security, economic issues, etc. Finally, medium-term forecasting stretches in a period from one week to one year and is usually applied for scheduling maintenance tasks.

Several scientific fields deal with signals that are governed by time-dependent relationships. In contrast to the typical machine learning (ML) task which is responsible for predicting continuous variables, namely regression, this kind of signals should be examined without violating their ordering. Time series forecasting is the most common term, under which such methods are categorized \cite{Hyndman2018}. 
Classical ML regression has been extensively applied to the task in the last years aiming to provide better results for both short-term and long-term scenarios, offering also increased robustness compared to pure statistical approaches. Following this direction, several types of ensemble methods have been recently considered in the context of time series forecasting, as they are described in \cite{Allende2017} and thoroughly investigated in \cite{DBLP:journals/asc/RibeiroC20}.

An equally important procedure for capturing the underlying components of generic time series signals is decomposition. Although the classic additive and multiplicative methods are encountered in many applications, there is a need for integrating more sophisticated strategies either with or without the former ones, in order to capture the distinct and heterogeneous components of time series. Toward this direction, a recently developed approach by Facebook, namely Prophet \cite{doi:10.1080/00031305.2017.1380080}, performs time series decomposition by employing generalized additive models (GAM) and Fourier series.

Despite the fact that Prophet has been used in several scientific papers as a competitor without actually managing to outperform various statistical and classical ML approaches (e.g., deaths caused by COVID-19 pandemic \cite{DBLP:conf/icccnt/KumarS20}), its striking success on more specific cases cannot be ignored (e.g., air pollution forecast \cite{10.1145/3355402.3355417}). This behavior is expected, if we consider that the original scope of the model was to forecast seasonal events that occur in the popular social media platform of Facebook. Hence, Prophet outperforms its competition when it comes to time-dependent signals that are governed by multiple seasonal effects. Such patterns are also encountered in RES, since their availability is affected by hourly, monthly, and seasonal variations. 

To the best of our knowledge, this is the first work to exploit the decomposition properties of Prophet, combined with a regression tree ensemble for short-term forecasting of energy generation by RES. In our approach, after collecting historical energy generation and weather forecasts, we constructed artificial features for capturing the additional seasonalities as exogenous regressors. Code and link for our approach and the new Greek RES datasets that we compiled are available online\footnote{\href{https://github.com/intelligence-csd-auth-gr/greek-solar-wind-energy-forecasting}{https://github.com/intelligence-csd-auth-gr/greek-solar-wind-energy-forecasting}}.

The rest of this paper is structured as follows: Section 2 presents recently published work involving time series forecasting in similar domains or using ensemble methods. Next, we provide a brief description of our examined dataset and the proposed feature engineering process. A narrative of the proposed algorithm is posed in Section 4, while our produced results along with some meaningful comments on the corresponding experimental procedure follow in Section 5. Finally, we sum up with the most important points of our work and pose future directions. 

%
%
\section{Related Work}

Regression ensembles for increasing the quality of time series forecasting have been explored in several approaches. Beyond the common methods of bagging and boosting, stacking was applied in \cite{DBLP:journals/asc/RibeiroC20}, where two datasets from the field of agribusiness were examined over short-term price forecasting. Models based on Random Forest (RF), as well as extreme gradient boosting and a stacking variant achieved the best error reduction rates. Four different metrics were examined: Mean Absolute Percentage Error (MAPE), Root Mean Square Error (RMSE), Mean Square Error(MSE) and Mean Absolute Error (MAE). These findings support the powerful modeling capabilities of tree-based ensemble regression. Instead of searching exhaustively for fixed-size combinations, the concept of negative correlation was employed in \cite{Allende2017} for selecting a subset of models among a predefined pool of candidates, without violating their bias-variance trade-off, but taking into consideration the covariance measure. 


A work that combines the Prophet model with Gaussian Process regression (GPreg)  \cite{DBLP:journals/chinaf/LiMPLY20}, managed to outperform auto-regressive integrated moving average (ARIMA) and wavelet-ARIMA approaches for predicting traffic in wireless networks. This phenomenon presents great fluctuations and heavily depends on user profile. They assume that the first approach can model better the long-range trend of the underlying signal, in contrast with the second, which tries to model the short-range as a multivariate problem through a kernel choice. An inverse discrete wavelet transformation was used to produce the final array of predictions. The train-test split was defined as 7-1 days at a time scale of 1 hour. Although the obtained results in terms of RMSE and MAPE were satisfactory, the main defect was the increased complexity of the GPreg model, a fact that may render this method infeasible when larger training datasets have to be tackled. 

In the domain of RES forecasting, the decomposition of the energy time series using wavelet transformation techniques (WTT) is a popular method to model the intermittent nature of solar and wind phenomena. In \cite{10.1016/j.renene.2019.03.020} the authors experiment with several models in their attempt to forecast photovoltaic (PV) power generation during sunny and cloudy days. They propose a hybrid model based on a Random Vector Functional Link neural network and a seasonal ARIMA (SARIMA) model to manage PV generation forecasting during cloudy days. The hybrid model employs the maximum overlap discrete wavelet transformation for time series decomposition, which distinguishes the original signal in a set of 5 series: a low frequency {\em approximation} and four high frequency {\em details}. The predictions of both models are combined for the final forecast. Although SARIMA models are known to handle seasonal variations, they do not account for multiple seasonalities \cite{multiple-seasonalities}, which are inherent in solar energy. On the other hand, dealing with multiple seasonalities is an important property of the Prophet model. 

In a similar fashion, in \cite{10.1016/j.asoc.2013.02.016} the authors present a hybrid model applied to short-term wind forecasting by combining WTT, seasonal adjustment method (SAM), and a radial basis function neural network (RBFNN). The proposed WTT–SAM–RBFNN approach can be described in 4 steps: first the WTT is used to de-noise the original time series by identifying the high and low frequency components, secondly the SAM decomposes the low frequency component, then the trend component is modeled using the RBFNN model and finally the hybrid prediction is achieved by combining the modeled trend and the seasonal indices. The proposed model is trained and tested using mean hourly wind speed data for one month in Wuwei and Minqin in China. Although the proposed model demonstrates high accuracy, the evaluation is restricted by the small dataset, in contrast to our work, which is applied to a dataset that consists of 4 years of hourly energy generation and weather forecasts.

\section{Dataset Description}
The goal of this work is to establish a system that can consistently and accurately predict renewable energy generation in Greece, using historical solar and wind energy generation data along with weather forecasts. The modeling and performance of such a system, depends on the forecasting constrains set by the real life applications. The examined scenario assumes a forecast horizon of one hour and data availability up to 48 hours. More specifically, previous information of wind and solar power generation is missing for the past 48 hours prior to the target period, while weather forecasts are available up to 24 hours after the target period. The historical RES generation data were collected by the European network of transmission system operators for electricity\footnote{\href{https://transparency.entsoe.eu/}{https://transparency.entsoe.eu}}, an online platform that operates as energy data aggregator for the 42 countries that participate in the centralized European energy market. Weather data were retrieved by the Storm Glass weather API\footnote{\href{https://stormglass.io/}{https://stormglass.io/}}.

Subsequently, four years of hourly data (2017-2020) were collected and used to create two datasets for the 1-step-ahead solar and wind energy forecasting tasks using: a) temporal features, b) endogenous properties of the signal, and c) exogenous weather variables according to their correlation with each energy type. Specifically, time-related features such as the hour of the day, day of the week, day of the month, day of the year, and month are treated as cyclical features encoded as polar coordinates. On the other hand, previous (lagged) energy observations are employed from 96 to 48 steps prior to the target, along with weather forecasts for 24 steps after and prior to the target. In addition, the statistical properties of the time series are captured as features by applying a rolling window and calculating the minimum, maximum, average, skewness, variance, and standard deviation of each energy type. Finally, the energy values were scaled using min-max normalization. 

\begin{figure}[h]
    \centering
    \begin{subfigure}{0.49\textwidth}
        \centering
        \includegraphics[width=\linewidth]{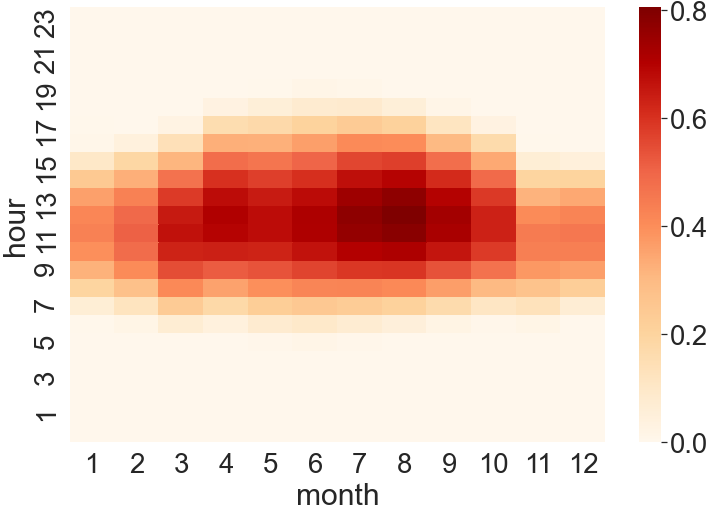} 
    \end{subfigure}
    \begin{subfigure}{0.49\textwidth}
        \centering
        \includegraphics[width=\linewidth]{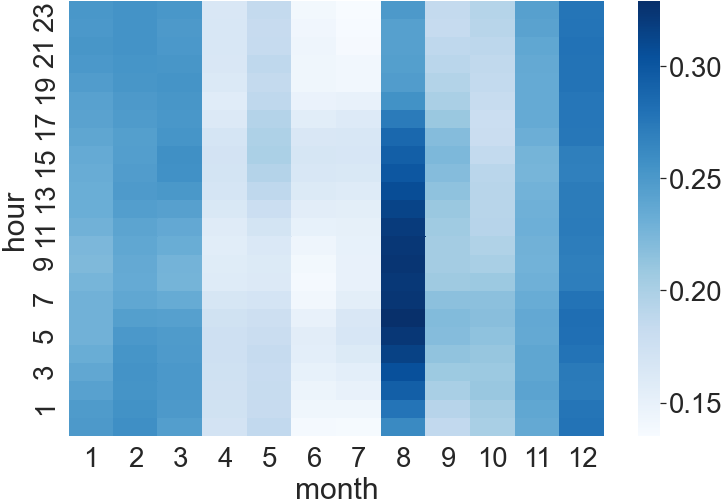} 
    \end{subfigure}
    \caption{Heat maps of scaled solar and wind energy generation for each hour of each month averaged over 4 years (2017-2020).}
    \label{fig:res-heatmap}
\end{figure}

Figure \ref{fig:res-heatmap} illustrates the production for each hour and month, demonstrating the various seasonalities that are inherent to the energy data. Solar energy reveals consistent daily and yearly patterns, whereas a yearly pattern is present in the wind heat map.

Each final dataset was formed after applying recursive feature elimination \cite{rfe}, using the feature importance as captured by ridge regression  \cite{DBLP:conf/nips/CortesM06} and MSE as the evaluation metric. Finally, out of the total 176 engineered features, 150 and 160 features were selected for solar and wind energy forecasting, respectively, avoiding the noisy and/or redundant features of the original higher dimensional space. Lastly, the first 3 years of data (2017-2019) were used for training and the last year (2020) for testing.

%
%
\section{Forecasting Models}
This section presents the various forecasting approaches that are used to compare and evaluate the proposed model during our experiments. We start with baseline models, continue with a classic ML model based on tree ensembles, as well as the state-of-the-art Prophet model. Finally, we conclude with the introduction of the proposed hybrid model.

\subsection{Baseline Models}

The baseline approach for modeling renewable energy generation in Greece is a persistence model \cite{https://doi.org/10.1002/swe.20040} based on the auto-correlation of the time series and the intuition that immediate past observations of a stochastic process will reflect the current observation better than older observations. This is formulated as:

\begin{equation}
    \hat{y}_t=y_{t-n}
\end{equation}

\noindent
where $n$ is the number of previous instances. Here, $n$ indicates the number of days in the past. Different time lags were used to generate three models for $t-2$, $t-7$ and $t-30$ days. 

\subsection{Machine Learning Models}
According to the literature, nonlinear models can be very effective in RES forecasting since they can capture the intermittent nature of the resources \cite{10.1016/j.renene.2016.12.095}. In this paper, we explore the performance of the Extra Trees (ExTs) \cite{DBLP:journals/ml/GeurtsEW06} regression model, which is an ensemble that fits multiple decision trees on randomized subsets of the training set. In regression, ExTs are applied by averaging the predictions of the decision trees. Its main difference with other popular ensemble models, such as RFs, lies in node splitting, where RFs depend on the calculation of a splitting criterion, whereas ExTs use a random splitting threshold. This random threshold decreases the computational burden, but usually makes the model depend on a greater number of estimators than RFs. In addition, ExTs use the whole feature space for training each estimator, while RFs use only a subset of features. The parameters for the ExTs model were selected using grid search parameter tuning, an exhaustive parameter optimization method that explores all possible configurations in a solution space. 

Furthermore, we employed the model of Prophet, which promises to be an automated, user friendly, easily tuned tool for analysts that can handle multiple seasonalities (Fig. \ref{fig:res-heatmap}). This feature is important in energy generation, since seasonal patterns in energy data display multi-period seasonalities (eg. daily, weekly and monthly). The Prophet is built as an additive regression model \cite{doi:10.1080/00031305.2017.1380080} in the form :

\begin{equation}
\label{prophet_decomposition}
    y(t) = g(t) + s(t) + h(t) + \epsilon_t
\end{equation}

\noindent
where $g(t)$ is the trend function, $s(t)$ the seasonal module that represents periodic changes, $h(t)$ the effect of holidays and $\epsilon_t$ is the error term. In addition, Prophet is customizable, allowing options such as the addition of custom seasonalities and the input of external regressors. By default, Prophet models seasonality using Fourier series which can be more efficient when it comes to periodic effects.

\subsection{Proposed hybrid model}

An additional property of the Prophet model is that it can act as a time series decomposition tool and offer insights into each examined problem. Here, we propose a hybrid model that exploits this property by removing the repeating temporal patterns from the energy time series, extrapolates the seasonality, and trains an ExTs regression model using the residual terms. The main intuition behind this modeling method is that the decomposable model can estimate the seasonal and trend components of the time series, while the ExTs model captures the non-linear patterns in the residual time series. Although a common method in time series decomposition literature is the employment of "approximations" and "details" extracted by  wavelet transformation \cite{10.1016/j.asoc.2013.02.016,10.1016/j.renene.2019.03.020}, Prophet uses a combination of GAM and Fourier series for the decomposition \cite{doi:10.1080/00031305.2017.1380080}. 

\begin{figure}[h]
    \centering
    \includegraphics[width=0.9\linewidth]{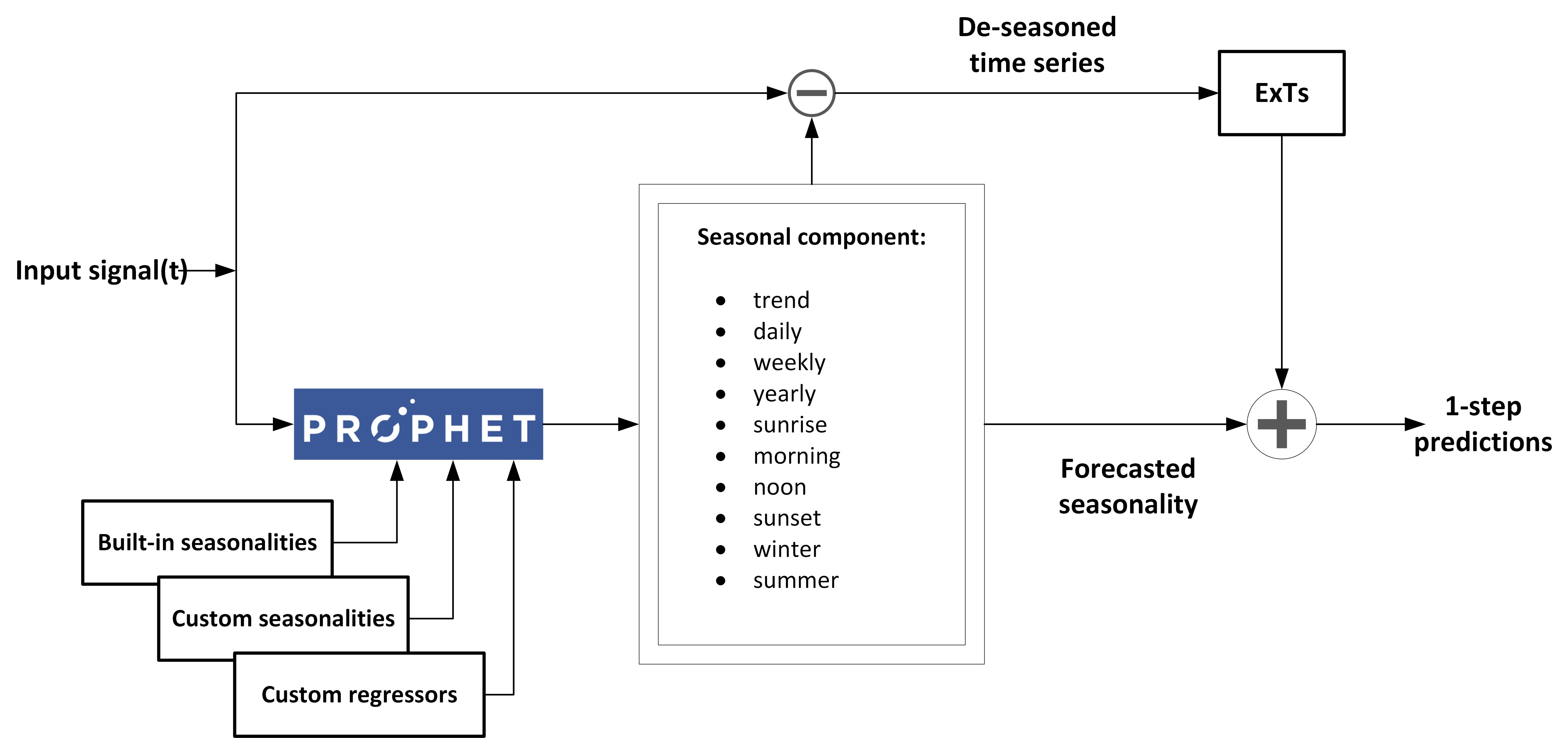} 
    \caption{Hybrid model using Prophet decomposition.}
    \label{fig:prophet-decomposition}
\end{figure}

The implementation of the hybrid model for the Greek RES generation forecasting system (see also Fig. \ref{fig:prophet-decomposition}), consists of four steps. First, we initialize Prophet by adding custom seasonalities and regressors, next the model is fitted to the training data and the seasonal component is calculated. Subsequently, the seasonal component is subtracted and the deseasoned time series is used as input to the ExTs model. The final forecast is produced by the residual forecasts plus the extrapolated seasonal patterns. 

\vspace*{-\baselineskip}

During the initialization of the Prophet model, apart from the default seasonalities, we introduced a biseasonal component to capture  seasonal variations (winter, summer) and a time of day component (sunrise, morning, noon, sunset, night). Furthermore, the following features were added as external regressors:
\begin{itemize}
    \item rolling statistics - mean, min, max, skewness, standard deviation, variance,
    \item previous energy values from 48 to 72 hours in the past,
    \item previous weather values from 48 to 72 hours in the past, and
    \item available forecasts for the next 24 hours of temperature, humidity, visibility and wind speed (solar energy modeling) and gust, wind speed (wind energy).
\end{itemize}

%
%
\section{Experimental Results}

This section presents the results of the forecasting experiments for each energy type and each model along with a commentary of the findings. To be consistent with the literature, the metrics of RMSE and MAE were used to measure the performance of the models \cite{DBLP:journals/asc/RibeiroC20}. In addition, the models are compared in terms of error intervals. Specifically, the interest lies in the percent of the absolute error that remains under 10\%, between 10\% and 15\%, and above 15\%. These intervals are of great interest to energy producers, revealing a sense of the forecasting confidence.

%




Table \ref{tab:metrics} shows the MAE and RMSE for all models and energy types. We first notice that the volatile wind signal is more challenging to forecast than the solar one, where the baselines are competitive. In addition, we notice the positive effect of the proposed feature engineering (FE) process, which strongly boosts the accuracy of all models in both energy types. Prophet is slightly better than ExTs in terms of RMSE, while ExTs are better in MAE, especially in terms of solar energy. Notably, the proposed hybrid model achieves the best results in both measures and energy types. 

\begin{table}[]
\centering
\caption{MAE and RMSE values for all models and both energy types.}
\label{tab:metrics}
\begin{tabular}{p{6cm} rrrr}
\toprule
\textbf{}                    & \multicolumn{2}{c}{\textbf{Solar}} & \multicolumn{2}{c}{\textbf{Wind}} \\ \hline
\textbf{Model} & \multicolumn{1}{l}{\textbf{MAE}} & \multicolumn{1}{l}{\textbf{RMSE}} & \multicolumn{1}{l}{\textbf{MAE}} & \multicolumn{1}{l}{\textbf{RMSE}} \\
\midrule
\textbf{Hybrid Prophet+ExTs}                 & \textbf{0.041}   & \textbf{0.067}  & \textbf{0.069}  & \textbf{0.088}  \\
Hybrid Prophet+ExTs w/o FE          & 0.077            & 0.117           & 0.106           & 0.136           \\
1-step-ahead ExTs Regression        & 0.045            & 0.081           & 0.081           & 0.11            \\
1-step-ahead ExTs Regression w/o FE & 0.146            & 0.195           & 0.107           & 0.144           \\
Prophet                             & 0.055            & 0.08            & 0.083           & 0.104           \\
Prophet w/o FE                      & 0.091            & 0.138           & 0.182           & 0.225           \\
Persistence t-2                     & 0.049            & 0.107           & 0.226           & 0.287           \\
Persistence t-7                     & 0.052            & 0.114           & 0.25            & 0.317           \\
Persistence t-30                    & 0.062            & 0.125           & 0.243           & 0.307           \\
\bottomrule
\end{tabular}
\end{table}

\begin{figure}[ht!]
    \centering
    \begin{subfigure}{1.\textwidth}
        \centering
        \includegraphics[width=\linewidth]{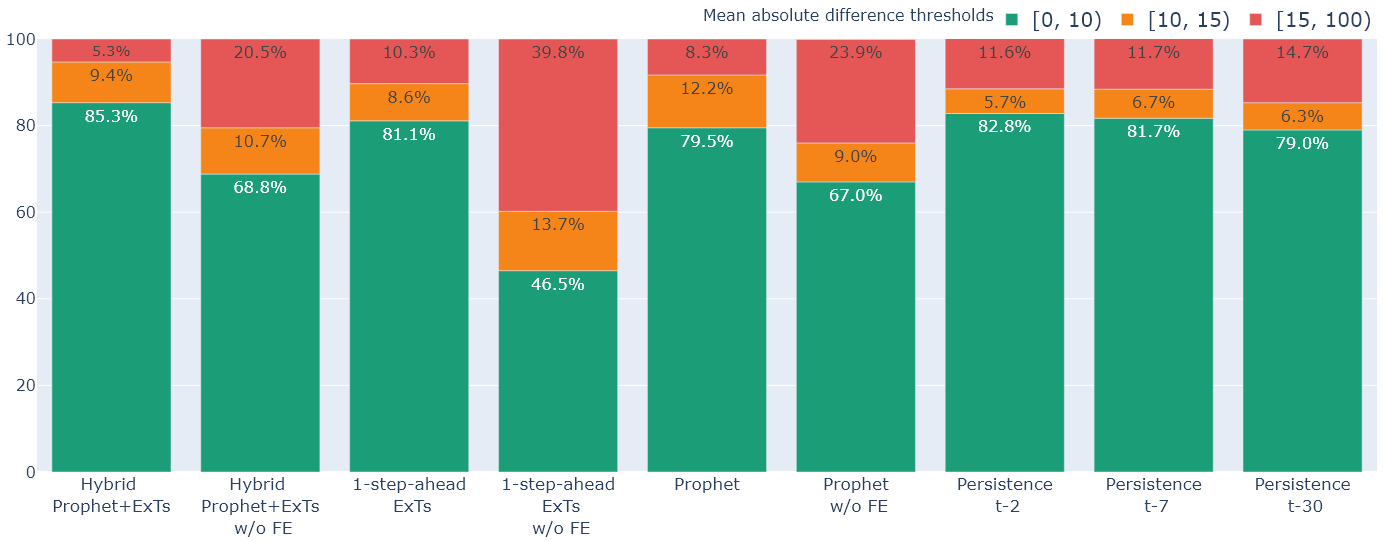} 
        \caption{Solar}
        \label{fig:solar-analysis}
    \end{subfigure}
    \begin{subfigure}{1.\textwidth}
       \centering
        \includegraphics[width=\linewidth]{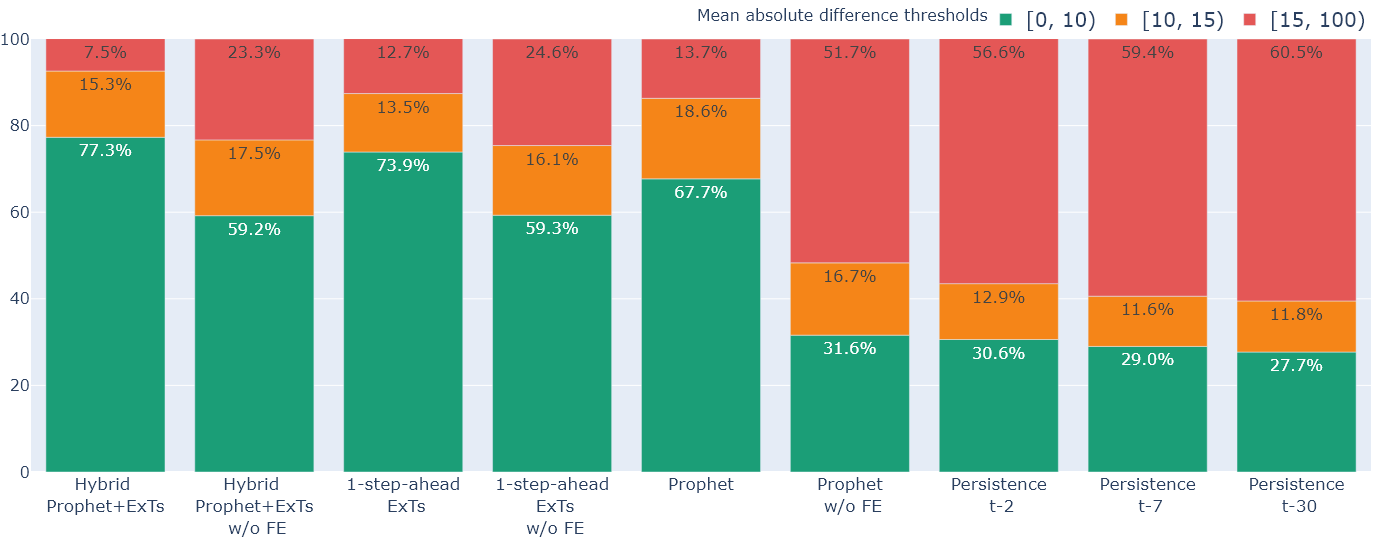} 
        \caption{Wind}
        \label{fig:wind-analysis}
    \end{subfigure}
    \caption{Distribution of absolute errors for each model and energy type.}
    \label{fig:energy-analysis}
\end{figure}

Fig. \ref{fig:energy-analysis} shows the distribution of the absolute errors for each model in each of the two energy types. The proposed hybrid approach is the most robust, achieving the highest percentage of errors below 10\% in both cases (85.3\% for solar and 77.5\% for wind energy) and the least amount of errors above the 15\% threshold, again in both cases (5.3\% for solar and 7.5\% for wind energy), following a short-tailed distribution. 

\begin{figure}[ht!]
    \centering
    \begin{subfigure}{\textwidth}
        \centering
        \includegraphics[width=\linewidth]{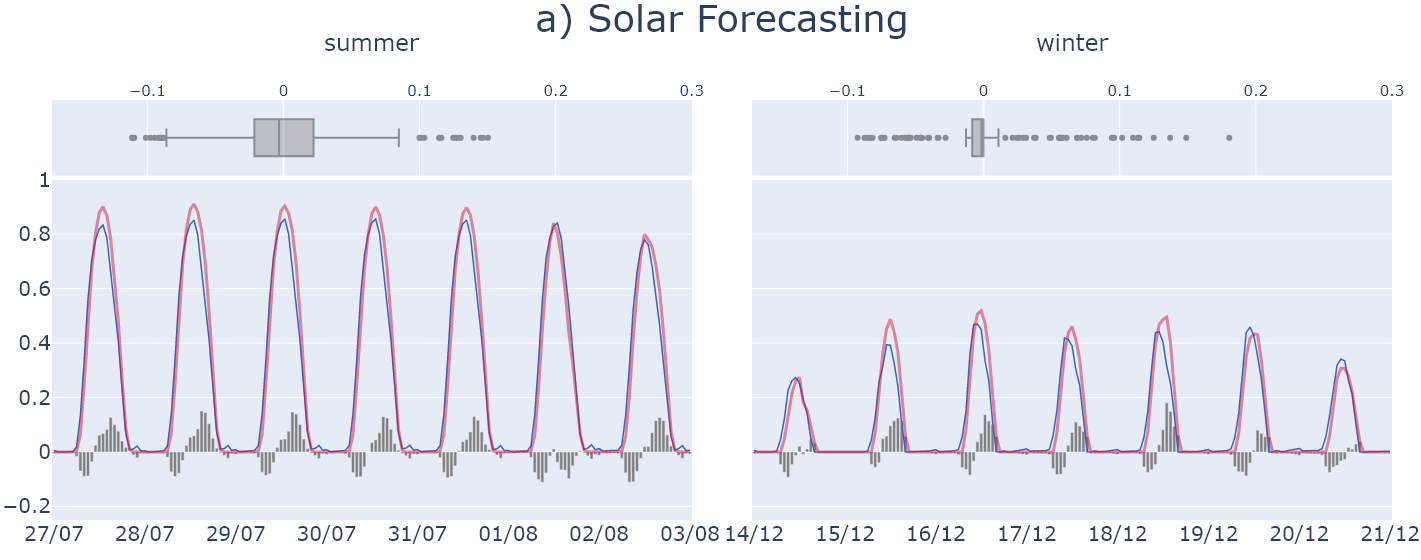}
        \label{fig:solar-forecasting}
    \end{subfigure}
    \begin{subfigure}{\textwidth}
        \centering
        \includegraphics[width=\linewidth]{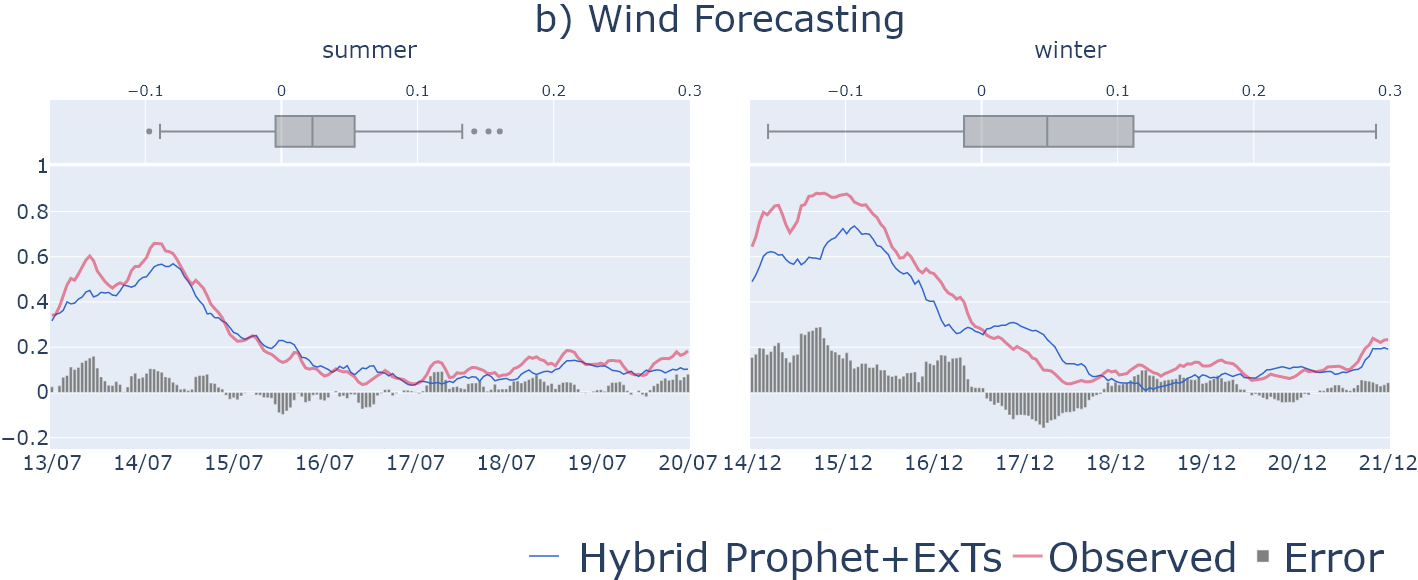}
        \label{fig:wind-forecasting}
    \end{subfigure}
    \caption{RES forecasts during a week in July and December.}
    \label{fig:results-forecasting}
\end{figure}

Lastly, Figure \ref{fig:results-forecasting} illustrates the actual and forecasted energy generation during a week in July and December. It becomes apparent that the consistent energy generation patterns are easily modeled by all models during summer, whereas during winter larger deviations can be observed.

\section{Conclusions and Future Work}

We presented an innovative combination of the Prophet model with non-linear tree-based ensembles, managing to achieve accurate and robust RES predictions. Among the main contributions of this work is the manufactured dataset that describes the solar and wind energy generation in Greece along with the proposed feature engineering and selection strategy. Moreover, the focal point of this paper is the utilization of Prophet as a time series decomposition model and its combination with ExTs to predict RES signals. 

Regarding future expansions of this work, one can experiment with the adoption of metrics that can select models with complementary predictive behavior and could operate as an asset for improving the performance of the proposed model \cite{Allende2017}. 
In many cases, energy forecasting predictions are needed for multiple steps in the future. Adapting the proposed model to accommodate multi-step forecasts is a topic for future research. Furthermore, different non-linear models can be combined with Prophet decomposition, especially neural network architectures such as LSTM networks that are often employed in forecasting tasks, or exploit them under meta-learning schemes for creating robust ensemble models \cite{DBLP:journals/access/VaiciukynasDKB21}.
Finally, more innovative parameter optimization techniques could be explored, such as genetic algorithms, instead of the naive grid search used here, which is tedious and computationally intensive \cite{DBLP:journals/gpem/RomanoLFM21}.

\subsubsection*{Acknowledgements.}
Part of this work is co‐financed by the European Regional Development Fund of the European Union and Greek national funds
through the Operational Program Competitiveness, Entrepreneurship and Innovation, under the call RESEARCH
– CREATE - INNOVATE (project code: T2EDK-03048).
%
%
\bibliographystyle{unsrt}
\bibliography{paper.bib}

\end{document}